\documentclass[conference]{IEEEtran}
\IEEEoverridecommandlockouts
\usepackage{cite}
\usepackage{amsmath,amssymb,amsfonts}
\usepackage{algorithmic}
\usepackage{graphicx}
\usepackage{textcomp}
\usepackage{xcolor}

\usepackage{hyperref}
\hypersetup{
    colorlinks=true,
    linkcolor=blue,
    filecolor=blue,      
    urlcolor=magenta,
    citecolor=magenta,
}

\usepackage{url}
\usepackage{booktabs}

\def\BibTeX{{\rm B\kern-.05em{\sc i\kern-.025em b}\kern-.08em
    T\kern-.1667em\lower.7ex\hbox{E}\kern-.125emX}}
\begin{document}

\title{PatchTrAD: A Patch-Based Transformer focusing on Patch-Wise Reconstruction Error for Time Series Anomaly Detection
\thanks{The authors acknowledge ANR – FRANCE (French National Research Agency) for its financial support of SHARP project ANR-23-IAS4-0003.}
}

\author{\IEEEauthorblockN{Samy-Melwan Vilhes}
\IEEEauthorblockA{\textit{INSA Rouen, Univ Rouen} \\
\textit{Normandie Univ, LITIS UR 4108}\\
F-76000 Rouen, France \\
samy-melwan.vilhes@insa-rouen.fr}
\and
\IEEEauthorblockN{Gilles Gasso}
\IEEEauthorblockA{\textit{INSA Rouen, Univ Rouen} \\
\textit{Normandie Univ, LITIS UR 4108}\\
F-76000 Rouen, France \\
gilles.gasso@insa-rouen.fr}
\and
\IEEEauthorblockN{Mokhtar Z. Alaya}
\IEEEauthorblockA{\textit{Univ de Technologie de Compiègne } \\
\textit{LMAC EA 222}\\
F-60203 Compiègne, France \\
alayaelm@utc.fr}
}

\maketitle

\begin{abstract}
Time series anomaly detection (TSAD) focuses on identifying whether observations in streaming data deviate significantly from normal patterns. With the prevalence of connected devices, %objects and IoT, 
anomaly detection on time series has become paramount, as it enables real-time monitoring and early detection of irregular behaviors across various application domains. In this work, we introduce PatchTrAD, a Patch-based Transformer model for time series anomaly detection. Our approach leverages a Transformer encoder along with the use of patches under a reconstruction-based framework for anomaly detection. %focused on minimizing reconstruction error. 
Empirical evaluations on multiple benchmark datasets show that PatchTrAD is on par, in terms of detection performance, with state-of-the-art deep learning models for anomaly detection while being time efficient during inference.

%We rigorously compare PatchTrAD with state-of-the-art deep learning models for anomaly detections on multiple benchmark datasets commonly used in the field. The results show that our model compete with previous state-of-the-arts methods while being time efficient during inference.

\end{abstract}


\section{Introduction}

%\noindent \textbf{Context.} 
Time series anomaly detection (TSAD) refers to the task of identifying whether new observations from a data stream significantly differ from expected normal patterns. 
%are normal or anomalous. 
Several real-world applications have been considered, including for instance,  industrial equipment status surveillance, intrusion detection or home monitoring. The even-increasing scale of sensing-technologies and their widespread in several application domains require efficient and accurate anomaly detection techniques to ensure security. The different types, dimensionality or properties of times series  has led to various anomaly detection methods for times series, including deep learning-based approaches \cite{doc, drocc, usad, donut, madgan, tranad, anotrans, dcdetector, patchad} %\rgg{mettre des ref ici}. 

%For instance, if we are monitoring a server composed of many CPUs, our goal is to detect anomalous utilization patterns that could potentially lead to a server crash. 
%As the Internet of Things continues to expand, the volume of data to process increases, along with the number of potential problems or attacks targeting these systems. Advancements in machine and deep learning offer effective solutions to manage this growing complexity. As we are considering real-time monitoring, these models must achieve high detection performance while remaining lightweight and ensuring fast inference, enabling rapid anomaly detection without compromising accuracy. \\
%\textbf{Existing approaches.} 
Many approaches for TSAD under unsupervised learning framework have been proposed. 
%Well-known machine learning-based models include Isolation Forest and One-Class SVM. 
Mainly, they can be categorized as reconstruction-based \cite{madgan, usad, tranad}, density-based or level set-based \cite{zong2018deep, doc, drocc}, contrastive learning \cite{anotrans, dcdetector, patchad} or prediction-based approaches \cite{wu2023timesnet, patchtst}. Reconstruction models aim to learn a latent representation of the data from which the original samples are reconstructed.  A high reconstruction error may be indicative of an anomaly.  Transformer encoder-decoder architectures \cite{tranad} are representatives of this category of algorithms with  promising detection performances. Density/level set-based methods typically perform density or level-set estimation from some latent representation of the time series and predict the likelihood or the score of new observations to be normal. Contrastive learning has been leveraged for TSAD and recently a   multiscale patch-based deep architecture \cite{patchad} that hinges on times series patch-mixing strategy to learn representation  adapted to anomaly detection has been introduced. Finally prediction-based approaches rely on recurrent cells such as LSTM  or Transformer-based deep architectures, including those using patches \cite{patchtst}, to train time-series forecasting models. An anomaly is deemed occurring when the forecasting error for given sequential new samples exceeds a certain threshold, indicating a significant change in the time series.

%In the context of deep learning models, predictive approaches such as basic LSTM networks or Transformer-based models, including those using patches \cite{patchtst}, are trained to forecast future values. When the forecasting error for a given observation exceeds a certain threshold, it may indicate a significant change in the time series, suggesting a potential anomaly. Reconstruction models \cite{madgan, usad, tranad} aim to learn a latent representation of the data and, from this representation, reconstruct the original observations. A high reconstruction error may indicate an anomaly. Transformer encoder-decoder architectures \cite{tranad} have shown promising results in this context.\\
%There are other types of models as well. Some leverage adversarial training to generate synthetic observations and train a basic classifier \cite{drocc}, while others use discrepancies in latent spaces to compute a score \cite{doc, anotrans, dcdetector, patchad}. \\
%\textbf{Objectives.} 
Building on the effectiveness of patch-based Transformer models for time series forecasting, %the complexity reduction 
the lightweight model achieved through patch construction and the efficiency of reconstruction-based methods for TSAD, we propose herein PatchTrAD a model that leverages these approaches to enhance anomaly detection.  %\\
%\textbf{Contributions.}
We show that our patch-based transformer model focusing on reconstruction error leads to state-of-the-art results on both univariate and multivariate time series while remaining fast during inference. %PatchTrAD is evaluated against multiple state-of-the-art deep learning-based models.
%We denote by $ p^+(x_{t+1} \mid x_{1:t})$ the distribution of the observation to test (\(x_{t+1}\)) under normal condition. Thus, our goal is to detect:
%\[
%k \quad \text{such that} \quad x_{t+k} \not\sim p^+(x_{t+k} \mid x_{1:t+k-1}).
%\]  
%In the previous example of the CPU of a server, the time series is univariate, meaning we consider only the CPU utilization, thus, \(x_t \in \mathbb{R}\). However, the time series can also be multivariate, where we consider not only the current CPU utilization but also the current CPU temperature and the temperature of the whole system. In this case, \(x_t \in \mathbb{R}^3\). More generally, if we consider \(M\) modalities, then \(x_t \in \mathbb{R}^M\).

\section{Related Works}
%In this work, we focus exclusively on deep learning based models for TSAD.
\subsection{Preliminary}

We formulate the problem as follows: let $ x_{1:t} = (x_1, x_2, \dots, x_t) $ denote a stream of data, where an observation at time \( t \) consists of \( M \) modalities ($x_t \in \mathbb{R}^{M}$; \quad $M = 1$ for univariate time series,  $M \geq 2$ for multivariate time series). The objective is to determine whether the next observation, \(x_{t+1}\) is normal or anomalous. In practice, one uses a sliding window of a predefined size \(w\) i.e., one relies on the the most recent \(w\) observations \(x_{t-w+1:t}\) to infer the normality of \(x_{t+1}\).

%When dealing with \(x_{t+1}\), the model will not consider the entire set of previous observations \(x_{1:t}\) for a computational reason, but only the most recent \(w\) observations \(x_{t-w+1:t}\).
%Most of the time models use a sliding window of a predefined size \(w\). When dealing with \(x_{t+1}\), the model will not consider the entire set of previous observations \(x_{1:t}\) for a computational reason, but only the most recent \(w\) observations \(x_{t-w+1:t}\).  

\subsection{Prediction error-based anomaly detection}

%The first methodology 
A category of techniques to detect anomaly in time series involves training prediction models. These models are trained using the 
%window 
samples \(x_{t-w+1:t}\) to predict \(x_{t+1}\). If the prediction error exceeds a predefined threshold,  \(x_{t+1}\) is deemed anomalous otherwise, it is considered normal. %The models we are interested in include a basic  LSTM-based model, a basic Transformer-based model, and PatchTST a multivariate time series forecasting Transformer-based model that utilizes patches along with the RevIN \cite{revin} invertible normalization technique.
Typical models include
LSTM-based model,  Transformer-based model \cite{transformer}. PatchTST is a Transformer-based model \cite{patchtst} that utilizes patches along with the RevIN \cite{revin} invertible normalization technique for handling multivariate time series.
%\end{itemize}

\subsection{Reconstruction error-based anomaly detection}

Another class of techniques are reconstruction models, which aim to reconstruct the input window. These models  commonly learn a latent representation space in an autoencoder manner based on windowed inputs \(x_{t-w+1:t+1}, \forall t \in [w,\ldots, T] \). At inference stage the model projects the input window \(x_{t-w+1:t+1}, \forall t > T \) into a latent space and reconstructs it back. Similarly to prediction models, if the reconstruction error exceeds a preset threshold, \(x_{t+1}\) is classified as anomalous. Example models are  LSTM-based autoencoder \cite{aelstm}, MAD-GAN \cite{madgan} a GAN-based multivariate time series model, USAD \cite{usad} a multivariate model with two autoencoders sharing the same encoder trained in adversarial way and TranAD \cite{tranad} a Transformer-based network that reconstructs the input window using a focus score-based self-conditioning.
%These models are trained taking as input the window along the test observation \(x_{t-w+1:t+1}\). Models project the input window into a latent space and try to reconstruct it. Similarly to predictive models, if the reconstruction error exceeds a predefined threshold, we consider \(x_{t+1}\) to be anomalous. The models we are interested in include a basic LSTM-based autoencoder, MAD-GAN \cite{madgan}: a GAN-based multivariate time series model, USAD \cite{usad}: a multivariate model with two autoencoders sharing the same encoder trained adversarially and TranAD \cite{tranad}: a Transformer-based network that reconstructs the input window using a focus-score.

\subsection{Other methods}

Other methods aim to determine a conformity score, for instance models relied on discrepancy in latent spaces (Anomaly Transformer \cite{anotrans}, DCdetecor \cite{dcdetector}, PatchAD \cite{patchad}). 
The Deep One-Class Classifier \cite{doc} simultaneously learns a lower-dimensional representation of normal windows and a one-class classifier that estimates the minimum volume data-enclosing hypersphere.  Test input windows lying outside the learned hypersphere is deemed abnormal.  The Deep Robust One-Class Classifier (DROCC) \cite{drocc} assumes that the typical training samples lie on a low dimensional locally linear manifold. DROCC employs a gradient ascent step to generate realistic anomalous samples, providing access to the negative class to enhance the anomaly detection. 
%\rgg{a ameliorer} \rvs{corrigé}

%The Deep One-Class Classifier \cite{doc} learns a lower-dimensional representation of normal windows and attempts to project them into a hypersphere of lower dimension. Since training is only done on normal windows, we aim to map anomalous windows outside the hypersphere. The Deep Robust One-Class Classifier (DROCC) \cite{drocc} uses a basic classifier to classify windows. The challenge, however, is that during training, we do not have access to anomalous windows. To address this, DROCC employs a gradient ascent step to generate anomalous windows that fool the classifier.

\section{PatchTrAD}

\begin{figure}[htbp]
\centerline{\includegraphics[width=1\linewidth]{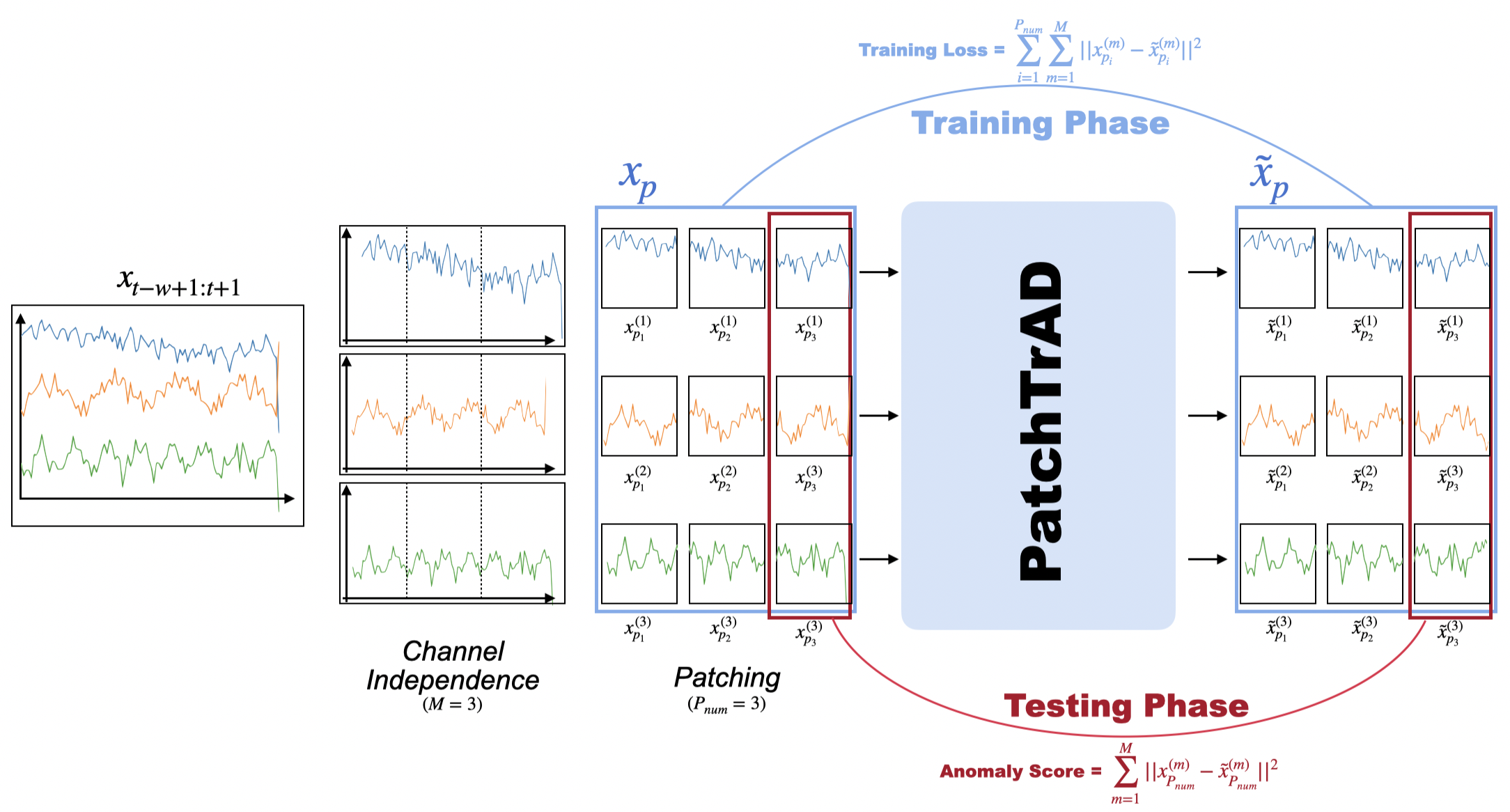}}
\caption{PatchTrAD Overview.}
\label{patchtrad}
\end{figure}

In this work, we propose \textbf{PatchTrAD}, a transformer-based reconstruction model that leverages patching techniques for TSAD and focuses on patch-wise reconstruction error. 
It is inspired from the time series forecasting model PatchTST \cite{patchtst}. 
An overview of PatchTrAD is detailed in~Fig. \ref{patchtrad}. 
% PatchTrAD (Fig. \ref{patchtrad}) %is highly inspired by  is inspired from the time series forecasting model PatchTST \cite{patchtst}. 
We adopt the concept of patches similar to the notion of tokens. Namely, patches/tokens are widely used in transformer architectures for vision and natural language processing (e.g. ViT~\cite{vit}, BERT~\cite{bert}), 
% and are crucial when working with local semantic information. 
and are crucial when dealing with local semantic information.
Patching for TSAD has been previously explored in \cite{dcdetector} and \cite{patchad}. 
We further incorporate the concept of channel independence, where each patch contains information from a single modality \(m \in \{1,\ldots,M\}\). 

\subsection{Patching}

The input of PatchTrAD is a window \( x_{t-w+1:t+1} \in \mathbb{R}^{M \times (w+1)} \). For a given  stream of a modality $m$, denoted as \( x^{(m)}_{t-w+1:t+1}\), its patch transformation is determined by the patch length \( P_{\text{len}} \) and the stride \( S \). Before patching, we pad the $m$-th stream by repeating  \( S \) times the last/test observation \( x^{(m)}_{t+1} \). Hence, patches can overlap and \( x^{(m)}_{t+1} \) belongs to the last patch. The number of patches is given by:
\[
P_{\text{num}} = \left\lfloor \frac{(w+1 - P_{\text{len}})}{S} \right\rfloor + 2.
\]
Thus, we transform the input window $x_{t-w+1:t+1} \in \mathbb{R}^{M \times (w+1)}$ into the tensor $x_p \in \mathbb{R}^{M \times P_{\text{num}} \times P_{\text{len}}}$.
%\[
%x_{t-w+1:t+1} \in \mathbb{R}^{M \times (w+1)} \to %x_p \in \mathbb{R}^{M \times P_{\text{num}} \times P_{\text{len}}}
%\]
We denote by \( x_p^{(m)} \in \mathbb{R}^{P_{\text{num}} \times P_{\text{len}}} \) the set of patches for the \( m \)-th modality extracted from $x_p$ and \( x_{p_i}^{(m)} \in \mathbb{R}^{P_{\text{len}}} \) represents the \( i \)-th patch for the \( m \)-th modality, with \( i \in \{1, \ldots, P_{\text{num}}\} \).

\subsection{Channel independence}

Channel independence, in contrast to channel dependency, refers to the scenario where each input patch contains information from only a single modality. Empirical evidence suggest that this setting does not harm performance \cite{channelind, patchtst}. In our approach, these input patches are fed into the same Transformer encoder, regardless of their modality. 
To make  our %future 
notations more readable, given a tensor \( y \in \mathbb{R}^{M \times N \times O} \) and a matrix \( W \in \mathbb{R}^{O \times D} \), the tensor-matrix product \( yW \) is computed by first flattening \( y \) into \( y \in \mathbb{R}^{(M N) \times O} \), performing the multiplication to obtain \( yW \in \mathbb{R}^{(M N) \times D} \), and then reshaping the result back to \( yW \in \mathbb{R}^{M \times N \times D} \). This clarification also applies to tensor-matrix addition.

\subsection{Vanilla transformer encoder layer}

Considering channel independence, we set a Vanilla Transformer Encoder \cite{transformer} that covers multiple layers of residual multi-head self-attention blocks with GELU activation, dropout, and batch normalization (omitted in the equations below). Note that the time dimension is represented by \( P_{\text{num}} \). 

In a first step, we project \(x_p\) using a learnable \(W_{\text{proj}} \in \mathbb{R}^{P_{\text{len}} \times D_{\text{model}}}\) and add a fixed positional encoding \(W_{\text{pe}} \in \mathbb{R}^{ P_{\text{num}} \times D_{\text{model}}}\), where \(D_{\text{model}}\) denotes the model dimension.
\[
\tilde{x}_p = x_p W_{\text{proj}} + W_{\text{pe}}  \in \mathbb{R}^{M \times P_{\text{num}} \times D_{\text{model}}}.
\]
The single-head attention block for one layer is defined by \(W_Q \in \mathbb{R}^{D_{\text{model}} \times D_k}\), \(W_K \in \mathbb{R}^{D_{\text{model}} \times D_k}\), \(W_V \in \mathbb{R}^{D_{\text{model}} \times D_v}\) and \(W_{\text{out}} \in \mathbb{R}^{D_v \times D_{\text{model}}}\) (only one head and one layer presented, with \(D_k\) and \(D_v\) hidden dimensions). Then:

\begin{align*}
    Q &= \tilde{x}_p W_Q \in \mathbb{R}^{M \times P_{\text{num}} \times D_k}, \\
    K &= \tilde{x}_p W_K \in \mathbb{R}^{M \times P_{\text{num}} \times D_k}, \\
    V &= \tilde{x}_p W_V \in \mathbb{R}^{M \times P_{\text{num}} \times D_v}, \\
    h &= \text{Softmax} \left( \frac{Q K^\top}{\sqrt{D_{\text{model}}}} \right) V \in \mathbb{R}^{M \times P_{\text{num}} \times D_v}, \\
    z &= h W_{\text{out}} \in \mathbb{R}^{M \times P_{\text{num}} \times D_{\text{model}}}.
\end{align*}

\subsection{Patch head}

From here, each modality has its own transformation. The patch head takes as input the output of the encoder \( z \in \mathbb{R}^{M \times P_{\text{num}} \times D_{\text{model}}} \). It projects each \(z^{(m)} \in \mathbb{R}^{P_{\text{num}} \times D_{\text{model}}}\) back to the patch length size  using \( M \) learnable linear functions \( W_{\text{out}}^m \in \mathbb{R}^{D_{\text{model}} \times P_{\text{len}}} \).
Therefore, we have:
\begin{align*}
& \Tilde{x}_p^{(m)} = z^{(m)} W_{\text{out}}^{m} \quad \in \mathbb{R}^{P_{\text{num}} \times P_{\text{len}}}, \\
& \Tilde{x}_p = \text{concat}(\Tilde{x}_p^{(1)}, \ldots, \Tilde{x}_p^{(M)}) \quad \in \mathbb{R}^{M \times P_{\text{num}} \times P_{\text{len}}}.
\end{align*}
A key difference from PatchTST \cite{patchtst} resides in this last layer: instead of flatten heads as %it was done 
in PatchTST, our approach  focuses solely on reconstructing the input patches.
% while PatchTST uses flatten heads to forecast future values, our approach avoids flatten heads and instead focuses solely on reconstructing the input patches.

\subsection{Training and detection}

Training PatchTST \cite{patchtst} leads to compute the MSE loss to compare forecasted values with the ground truth. However PatchTrAD is designed to accurately reconstruct the entire input patch \(x_p\). Thus, the training loss function we consider is the sum squared error between \( x_p \) and its reconstruction \( \tilde{x}_p \).
\[
\text{training loss} = \sum_{i=1}^{P_{\text{num}}} \sum_{m=1}^M \lvert \lvert x_{p_i}^{(m)} - \tilde{x}_{p_i}^{(m)} \lvert \lvert^2.
\]
The patching setting of PatchTrAD ensures that the test observation \( x_{t+1} \) is always included in the final patch. During detection phase, the anomaly score is computed through  the reconstruction error of the last patch \(x_{P_{\text{num}}}\), as it focuses on the final observation—the one under evaluation:
\[
\text{anomaly score} = \sum_{m=1}^M \lvert \lvert x_{P_{\text{num}}}^{(m)} - \tilde{x}_{P_{\text{num}}}^{(m)} \lvert \lvert^2.
\]
A higher anomaly score implies a greater likelihood that the test observation is anomalous according to our model.

\section{Experiments}

\subsection{Datasets}

To compare PatchTrAD to the state-of-the-art models, we conduct experiments on several datasets, being univariate and multivariate time series. For each dataset, the training set is only composed of normal observations while the test set contains normal and anomalous observations. \\
In the univariate case, we consider two datasets: \textit{NYC taxi demand} dataset (0.11\% anomalies in test set, \(M\)=1) and \textit{CPU usage} data from an Amazon's server in a datacenter  (0.15\%, \(M\)=1). Both datasets are taken from Numenta Anomaly Benchmark (NAB) \cite{nab}. For the multivariate case, we consider several datasets: \textit{Secure Water Treatment (SWaT) Dataset}\footnote{Credited to iTrust, Centre for Research in Cyber Security, Singapore University of Technology and Design.} (12\%, \(M\)=51); \textit{Server Machine Dataset}\footnote{Credited to the Tsinghua Netman Lab: \texttt{\url{https://github.com/NetManAIOps/OmniAnomaly}}} (4\%, \(M\)=38); and two NASA datasets: \textit{Mars Science Laboratory (MSL)} satellite dataset (10\%, \(M\)=55) and \textit{Soil Moisture Active Passive (SMAP)} rover dataset\footnote{Credited to the NASA Jet Propulsion Laboratory: \texttt{\url{https://github.com/khundman/telemanom}}} (13\%, \(M\)=25). SMD, SMAP and MSL are composed of several sub-datasets, We evaluate each model on every sub-dataset and average the performance. \\

\begin{table*}[htbp]
\caption{ROC-AUC scores (\textbf{bold}: first, \underline{underline}: second, \textit{italic}: third)}
\label{scores}
\begin{center}
\begin{tabular}{l|rrrrrr|r|l}
\toprule
\textbf{Dataset} & NYC-Taxi & EC2 & MSL & SWaT & SMAP & SMD & Mean & \textbf{Rank} \\
\midrule
\textbf{Model} & & & & \textbf{ROC-AUC} & & & &  \\
\midrule
DC-Detector & 0.498 & 0.827 & 0.536 & 0.435 & 0.560 & 0.530 & 0.564 & 11.8 \\
DROCC & 0.529 & 0.886 & 0.531 & 0.751 & 0.569 & 0.638 & 0.651 & 11.0\\
MADGAN & \textit{0.782} & 0.011 & 0.499 & 0.791 & 0.544 & 0.708 & 0.556 & 10.0 \\
USAD & 0.675 & \textit{0.977} & \underline{0.622} & 0.814 & 0.448 & 0.638 & 0.696 & 8.8 \\
PatchTST-rev$^{\mathrm{a}}$ & 0.552 & \textbf{0.999} & 0.562 & 0.233 & 0.498 & \textit{0.873} & 0.620 & 8.7 \\
DOC & 0.704 & 0.804 & 0.603 & 0.404 & 0.583 & 0.766 & 0.644 & 8.5 \\
LSTM-rev$^{\mathrm{a}}$ & 0.646 & \underline{0.998} & 0.598 & 0.238 & 0.520 & 0.858 & 0.643  & 8.4 \\
AnomalyTransformer & 0.491 & 0.994 & \textit{0.609} & 0.819 & \underline{0.637} & 0.678 & 0.705 & 7.7 \\
LSTM & 0.511 & \textbf{0.999} & 0.582 & \textit{0.842} & 0.604 & 0.833 & 0.729 & 6.5 \\
AE-LSTM & 0.664 & \underline{0.998} & 0.589 & 0.840 & 0.614 & 0.828 & \textit{0.756} & 6.0  \\
PatchTST & 0.696 & \textbf{0.999} & 0.560 & \underline{0.843} & 0.514 & \underline{0.882} & 0.749 & 5.5 \\
TranAD & 0.551 & 0.967 & \underline{0.622} & 0.815 & \textbf{0.668} & \textbf{0.884} & 0.751 & \textit{5.2} \\
PatchAD & \textbf{0.972} & \underline{0.998} & \textbf{0.625} & 0.822 & \textit{0.630} & 0.818 & \underline{0.811} & \underline{4.1}\\
\textbf{PatchTrAD} (our) & \underline{0.922} & \textbf{0.999} & \underline{0.622} & \textbf{0.845} & 0.629 & 0.869 & \textbf{0.814} & \textbf{2.8} \\
\bottomrule
{$^{\mathrm{a}}$Revin normalization applied.}
\end{tabular}
\end{center}
\end{table*}

\subsection{Evaluation method}

Most prior works on deep learning for TSAD do not rely on the ROC-AUC score, despite its effectiveness in comparing models on various datasets with different class imbalance \cite{paperbiosurmonbureau}.
%\begin{quote}
%    \textit{The ROC-AUC provides a robust estimation and ranking of classifier performance across %different class imbalances, while PR-AUC changes drastically} \cite{paperbiosurmonbureau}.
%\end{quote}
Instead, they primarily report F1-Score, Precision, and Recall, 
after using \textit{Point Adjustment} (PA) method used for the first time in \cite{donut}.
% However, these metrics require determining a threshold, but this choice depends on the application, as some problems tolerate false positives more than others. 
However, these metrics require setting a threshold,  but this choice depends on the application.

It is worth to note that PA algorithm modifies the model’s detections using ground-truth labels before evaluation. Specifically, it considers an entire anomaly period as correctly detected if the model identifies at least one anomaly within that period. PA improves significantly the model's performance, to the point where even a random model can achieve strong detection performance (measured for instance by a F1 score). %really strong scores (such as F1). 
A detailed study challenging this method %was presented 
is in \cite{cheat}. % For our evaluation, we then do not employ PA method.

% Consequently, we do not employ PA method. This algorithm modifies the model’s detections using ground-truth labels before evaluation. Specifically, it considers an entire anomaly period as correctly detected if the model identifies at least one anomaly within that period. This algorithm significantly improves the model's performance, to the point where even a random model can achieve really strong scores (such as F1). A study challenging this method was conducted in \cite{cheat}.

% Since our focus is on real-time anomaly detection, we do not consider this evaluation metric pertinent. 
To ensure a fair and easily interpretable comparison, we rely solely on ROC-AUC score without applying PA. This evaluation scheme eliminates the need to determine a threshold for model comparison, which is implicitly handled within ROC-AUC metric. 

\subsection{Pre-processing and hyperparameters}

We normalize each modality of the time series using statistics computed from the training set. This ensures consistency across all models, as they share the same preprocessing steps. Considering hyperparameters, we use the same batch size and window size for each model, adjusting them based on the dataset. For PatchTrAD, we use a patch size of $8$ and a stride of $6$ each time. Thus, patches overlap and, by construction, the observation to test is on the last patch. We replicate original implementations from the authors' GitHub repositories. When necessary, we make slight modifications to the models dimension to ensure they fit within a single GPU (NVIDIA RTX 2000 Ada Generation Laptop GPU). This adjustment is crucial, as we focus on real-time applications where very large models may be impractical for continuous deployment in production environments.

\subsection{Results}

\begin{figure}[h]
\centerline{\includegraphics[width=0.9\linewidth]{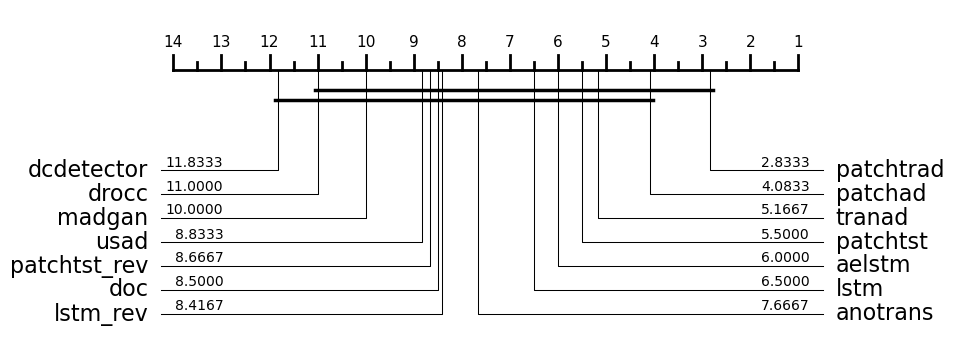}}
\caption{Critical difference diagrams for AUC scores using the post-hoc Nemenyi test with \( \alpha=5\% \), where better-ranked methods appear on the upper right.}
\label{nemenyi}
\end{figure}

As shown in Table \ref{scores}, PatchTrAD competes with the top-performing models. It achieves the best performance according to its rank and overeall mean performance. Both PatchTrAD and PatchAD leverage the patching technique. However, PatchTrAD is a reconstruction-based model using attention, whereas PatchAD is a discrepancy-based model without attention. Another top competitor is TranAD, which is also a reconstruction-based model with attention but does not incorporate patching. TranAD excels on multivariate datasets but performs less effectively than PatchTrAD and PatchAD on univariate datasets. \\
We rank all methods using the post-hoc Nemenyi test \cite{demsar}. The diagram in Fig. \ref{nemenyi} serves not as a definitive conclusion but as one from several ways to describe the performance of the predictors.
According to this test, PatchTrAD ranks first, followed by PatchAD in second place. TranAD achieves a better mean rank than the LSTM-based AutoEncoder (while TranAD achieves a lower overall mean AUC). Additionally, we identify distinct groups of models with significantly different performance levels (bold lines). The first group includes all models except PatchTrAD, indicating no significant difference among them. Since PatchTrAD ranks first, this suggests that it is the best-performing model according to this test. Conversely, the second group consists of all models except DCDetector, the worst-performing model. This suggests that all models perform similarly, except for DCDetector, which is noticeably less effective. 
The implementation of PatchTrAD is available at: \texttt{\url{https://github.com/vilhess/PatchTrAD}}

\subsection{Inference-time computation}

\begin{figure}[htbp]
\centerline{\includegraphics[width=0.7\linewidth]{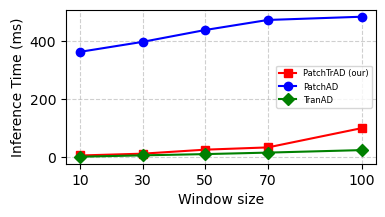}}
\caption{Inference-time Computation based on SWaT Dataset configuration for various window sizes, with a batch size of $128.$}
\label{infspeed}
\end{figure}

As previously concluded, three models stand out: PatchTrAD (ours), PatchAD, and TranAD. Since we focus on real-time anomaly detection, the models under consideration should be both fast and efficient during inference.
In Fig. \ref{infspeed}, we depict inference times of these models according to $w$, the size of the time window. As it can be noticed, PatchAD is by far the most time-consuming and PatchTrAD is more efficient than PatchAD. 
However, PatchTrAD is still behind TranAD, and this gap becomes  more %notable 
noticeable as the window size increases.
% When comparing these models in terms of inference time (Fig. \ref{infspeed}), PatchAD is by far the most time-consuming. PatchTrAD is more efficient than PatchAD but is still behind TranAD, with this gap becoming more pronounced as the window size increases.

\section{Conclusion}
We introduced PatchTrAD, a transformer-based model leveraging patches for anomaly detection focusing on reconstruction error. This model competes with state-of-the-art approaches and is 
almost 3 times faster than the best-competitor model PatchAD. 
It performs well across diverse datasets and remains efficient during inference, making it suitable to a wide range of univariate and multivariate time series.
We hence believe that PatchTrAD might be strong potential for further industrial TSAD problems. Future work could explore pretraining the transformer encoder on a diverse range of time series, followed by fine-tuning the patch head for each new time series, as this approach would improve generalization and enable efficient transfer learning.
% demonstrates more than thrice the time-efficiency of the best-competitor model while achieving better results. 
% For industrial deployment, PatchTrAD shows strong potential. 
% It performs well across diverse industrial datasets and remains efficient during inference, making it adaptable to a wide range of industrial time series, both univariate and multivariate.  

\section{Appendices}
%\subsection{Datasets Statistics}

%In \ref{tab:dataset_stats}, we present the statistics for each dataset. These statistics depend on our own implementation, including how we handle missing values and other preprocessing steps. As a result, they may differ from those reported in previous papers. Some prior works provide only post-processed datasets stored in cloud repositories, without detailing their preprocessing methods. In contrast, our implementation ensures full transparency, allowing a clear understanding of how the datasets are managed.

%\begin{table}[h]
%    \centering
%    \caption{Datasets Statistics}
%    \begin{tabular}{lcccc}
%        \toprule
%        Dataset & Dimension & Train Size & Test Size & Anomalies (\%) \\
%        \midrule
%        NYC Taxi & 1 & 5570 & 4750 & 0.11 \\
%        EC2 & 1 & 1984 & 2049 & 0.15 \\
%        SWAT & 51  & 495000  & 449919 &  12.13\\
%        SMD & 38 & 708405 & 708420 & 4.16 \\
%        MSL & 55 & 58317 & 73729 & 10.48 \\
%        SMAP & 25 & 140825 & 444035 & 12.85 \\
%        \bottomrule
%    \end{tabular}
%    \label{tab:dataset_stats}
%\end{table}

\subsection{Ablation study: patch size and stride impact}
PatchTrAD's architecture is determined by the patch size and stride, which together define the number of patches. In this section, we analyze how these parameters impact the final score by evaluating the model exclusively on NYC Taxi Demand and SWaT datasets.

\begin{table}[htbp]
\caption{ROC-AUC of PatchTrAD with Varying Strides and Patch Sizes\\
(\textbf{bold}: first, \underline{underline}: second)}
\label{abl}
\begin{center}
\begin{tabular}{l@{\hskip 0.2pt}r|rr}
\toprule
\textbf{Dataset} & & NYC Taxi & SWaT  \\
& & (\(w=32\)) & (\(w=100\))  \\
\midrule
\(P_{\text{len}}\) & \(S\) & \multicolumn{2}{c}{\textbf{ROC-AUC}} \\
\midrule
3  & 3  & 0.776 & 0.839 \\
5  & 3  & 0.904 & 0.839 \\
5  & 5  & 0.832 & 0.839 \\
6  & 6  & 0.872 & 0.842 \\
8  & 3  & 0.838 & \textbf{0.846} \\
8  & 5  & 0.801 & 0.844 \\
8  & 6  & \textbf{0.922} & \underline{0.845} \\
8  & 8  & \underline{0.917} & \underline{0.845} \\
16 & 12 & 0.536 & 0.821 \\
16 & 16 & 0.801 & 0.820 \\
28 & 22 & 0.890 & 0.822 \\
28 & 28 & 0.544 & 0.823 \\
32 & 28 & 0.549 & 0.829 \\
32 & 32 & 0.568 & 0.825 \\
\bottomrule
\end{tabular}
\end{center}
\end{table}

%\begin{table*}[htbp]
%\caption{ROC-AUC of PatchTrAD with Varying Strides and Patch Sizes\\
%(\textbf{bold}: first, \underline{underline}: second)}
%\label{abl}
%\begin{center}
%\begin{tabular}{l|rrrrrrrrrrrrrr}
%\toprule
%\textbf{Metric} & 3-3 & 5-3 & 5-5 & 6-6 & 8-3 & 8-5 & 8-6 & 8-8 & 16-12 %& 16-16 & 28-22 & 28-28 & 32-28 & 32-32 \\
%\midrule
%NYC Taxi (\(w=32\)) & 0.776 & 0.904 & 0.832 & 0.872 & 0.838 & 0.801 & %\textbf{0.922} & \underline{0.917} & 0.536 & 0.801 & 0.890 & 0.544 & %0.549 & 0.568 \\
%SWaT (\(w=100\)) & 0.839 & 0.839 & 0.839 & 0.842 & \textbf{0.846} & %0.844 & \underline{0.845} & \underline{0.845} & 0.821 & 0.820 & 0.822 & %0.823 & 0.829 & 0.825 \\
%\bottomrule
%\end{tabular}
%\end{center}
%\end{table*}

We observe in Table \ref{abl} that for PatchTrAD to perform optimally, it's crucial to find the right balance. If the patch size and stride are too large, performance decreases. Conversely, if they are too small, the model does not achieve its best results. Based on our experiments, a patch length of $8$ and a stride of $6$ yield the best detection performances. We do not consider strides greater than the patch length, as this would result in not considering all observations within the window.

%\subsection{Ranking of Methods} % refaire cette partie
%To conclude, we rank all methods using the post-hoc Nemenyi test \cite{demsar}. The diagram serves not as a definitive conclusion but as one of several ways to describe the performance of the estimators.
%\begin{figure}[h]
%\centerline{\includegraphics[width=0.9\linewidth]{cd.png}}
%\caption{Critical difference diagrams for AUC scores using the post-hoc Nemenyi test, where higher-ranked methods appear on the right.}
%\label{nemenyi}
%\end{figure}
%According to this test, PatchTrAD still ranks first. PatchAD comes in second. TranAD achieves a better rank than PatchAD (while TranAD achieves a lower overall mean AUC). Furthermore, we observe distinct groups of models that differ significantly from the rest (bold lines). In the first group setting, the three best models—PatchTrAD, TranAD, and PatchAD—perform significantly better than the others. Conversely, in the second group setting, the worst model, DCDetector, is significantly worse than the rest.

\bibliographystyle{ieeetr}
\bibliography{biblio}

\begin{thebibliography}{10}

\bibitem{doc}
L.~Ruff, R.~Vandermeulen, N.~Goernitz, L.~Deecke, S.~A. Siddiqui, A.~Binder, E.~M{\"u}ller, and M.~Kloft, ``Deep one-class classification,'' in {\em Proceedings of the 35th International Conference on Machine Learning} (J.~Dy and A.~Krause, eds.), vol.~80 of {\em Proceedings of Machine Learning Research}, pp.~4393--4402, PMLR, 10--15 Jul 2018.

\bibitem{drocc}
S.~Goyal, A.~Raghunathan, M.~Jain, H.~Simhadri, and P.~Jain, ``Drocc: deep robust one-class classification,'' in {\em Proceedings of the 37th International Conference on Machine Learning}, ICML'20, JMLR.org, 2020.

\bibitem{usad}
J.~Audibert, P.~Michiardi, F.~Guyard, S.~Marti, and M.~A. Zuluaga, ``Usad: Unsupervised anomaly detection on multivariate time series,'' in {\em Proceedings of the 26th ACM SIGKDD International Conference on Knowledge Discovery \& Data Mining}, KDD '20, (New York, NY, USA), p.~3395–3404, Association for Computing Machinery, 2020.

\bibitem{donut}
H.~Xu, W.~Chen, N.~Zhao, Z.~Li, J.~Bu, Z.~Li, Y.~Liu, Y.~Zhao, D.~Pei, Y.~Feng, J.~Chen, Z.~Wang, and H.~Qiao, ``Unsupervised anomaly detection via variational auto-encoder for seasonal kpis in web applications,'' in {\em Proceedings of the 2018 World Wide Web Conference}, WWW '18, (Republic and Canton of Geneva, CHE), p.~187–196, International World Wide Web Conferences Steering Committee, 2018.

\bibitem{madgan}
D.~Li, D.~Chen, B.~Jin, L.~Shi, J.~Goh, and S.-K. Ng, ``Mad-gan: Multivariate anomaly detection for time series data with generative adversarial networks,'' in {\em Artificial Neural Networks and Machine Learning -- ICANN 2019: Text and Time Series} (I.~V. Tetko, V.~K{\r{u}}rkov{\'a}, P.~Karpov, and F.~Theis, eds.), (Cham).

\bibitem{tranad}
S.~Tuli, G.~Casale, and N.~R. Jennings, ``Tranad: deep transformer networks for anomaly detection in multivariate time series data,'' {\em Proc. VLDB Endow.}, vol.~15, p.~1201–1214, Feb. 2022.

\bibitem{anotrans}
J.~Xu, H.~Wu, J.~Wang, and M.~Long, ``Anomaly transformer: Time series anomaly detection with association discrepancy,'' in {\em International Conference on Learning Representations}, 2022.

\bibitem{dcdetector}
Y.~Yang, C.~Zhang, T.~Zhou, Q.~Wen, and L.~Sun, ``Dcdetector: Dual attention contrastive representation learning for time series anomaly detection,'' in {\em Proceedings of the 29th ACM SIGKDD Conference on Knowledge Discovery and Data Mining}, KDD '23, (New York, NY, USA), p.~3033–3045, Association for Computing Machinery, 2023.

\bibitem{patchad}
Z.~Zhong, Z.~Yu, Y.~Yang, W.~Wang, and K.~Yang, ``Patchad: A lightweight patch-based mlp-mixer for time series anomaly detection,'' 2024.

\bibitem{zong2018deep}
B.~Zong, Q.~Song, M.~R. Min, W.~Cheng, C.~Lumezanu, D.~Cho, and H.~Chen, ``Deep autoencoding gaussian mixture model for unsupervised anomaly detection,'' in {\em International Conference on Learning Representations}, 2018.

\bibitem{wu2023timesnet}
H.~Wu, T.~Hu, Y.~Liu, H.~Zhou, J.~Wang, and M.~Long, ``Timesnet: Temporal 2d-variation modeling for general time series analysis,'' in {\em The Eleventh International Conference on Learning Representations}, 2023.

\bibitem{patchtst}
Y.~Nie, N.~H. Nguyen, P.~Sinthong, and J.~Kalagnanam, ``A time series is worth 64 words: Long-term forecasting with transformers,'' in {\em The Eleventh International Conference on Learning Representations}, 2023.

\bibitem{transformer}
A.~Vaswani, N.~Shazeer, N.~Parmar, J.~Uszkoreit, L.~Jones, A.~N. Gomez, L.~Kaiser, and I.~Polosukhin, ``Attention is all you need,'' in {\em Proceedings of the 31st International Conference on Neural Information Processing Systems}, NIPS'17, (Red Hook, NY, USA), p.~6000–6010, Curran Associates Inc., 2017.

\bibitem{revin}
T.~Kim, J.~Kim, Y.~Tae, C.~Park, J.-H. Choi, and J.~Choo, ``Reversible instance normalization for accurate time-series forecasting against distribution shift,'' in {\em International Conference on Learning Representations}, 2022.

\bibitem{aelstm}
O.~I. Provotar, Y.~M. Linder, and M.~M. Veres, ``Unsupervised anomaly detection in time series using lstm-based autoencoders,'' in {\em 2019 IEEE International Conference on Advanced Trends in Information Theory (ATIT)}, pp.~513--517, 2019.

\bibitem{vit}
A.~Dosovitskiy, L.~Beyer, A.~Kolesnikov, D.~Weissenborn, X.~Zhai, T.~Unterthiner, M.~Dehghani, M.~Minderer, G.~Heigold, S.~Gelly, J.~Uszkoreit, and N.~Houlsby, ``An image is worth 16x16 words: Transformers for image recognition at scale,'' in {\em International Conference on Learning Representations}, 2021.

\bibitem{bert}
J.~Devlin, M.-W. Chang, K.~Lee, and K.~Toutanova, ``{BERT}: Pre-training of deep bidirectional transformers for language understanding,'' in {\em NAACL} (J.~Burstein, C.~Doran, and T.~Solorio, eds.), (Minneapolis, Minnesota), pp.~4171--4186, Association for Computational Linguistics, June 2019.

\bibitem{channelind}
L.~Han, H.-J. Ye, and D.-C. Zhan, ``The capacity and robustness trade-off: Revisiting the channel independent strategy for multivariate time series forecasting,'' 2023.

\bibitem{nab}
A.~Lavin and S.~Ahmad, ``Evaluating real-time anomaly detection algorithms -- the numenta anomaly benchmark,'' in {\em 2015 IEEE 14th International Conference on Machine Learning and Applications (ICMLA)}, IEEE, Dec. 2015.

\bibitem{paperbiosurmonbureau}
E.~Richardson, R.~Trevizani, J.~A. Greenbaum, H.~Carter, M.~Nielsen, and B.~Peters, ``The receiver operating characteristic curve accurately assesses imbalanced datasets,'' {\em Patterns}, vol.~5, no.~6, p.~100994, 2024.

\bibitem{cheat}
K.~Siwon, K.~Choi, H.-S. Choi, B.~Lee, and S.~Yoon, ``Towards a rigorous evaluation of time-series anomaly detection,'' {\em Proceedings of the AAAI Conference on Artificial Intelligence}, vol.~36, pp.~7194--7201, 06 2022.

\bibitem{demsar}
J.~Dem{\v{s}}ar, ``Statistical comparisons of classifiers over multiple data sets,'' {\em Journal of Machine Learning Research}, vol.~7, no.~1, pp.~1--30, 2006.

\end{thebibliography}

\end{document}